\documentclass{article}
\usepackage{spconf,amsmath,graphicx,color}
\usepackage{placeins}
\usepackage{amssymb}
\usepackage{url}
\usepackage{bbm}
\usepackage{subcaption}
\usepackage{caption}
\usepackage{subcaption}

\definecolor{pink}{RGB}{219, 48, 122}
\newcommand{\marcely}[1]{\textcolor{blue}{#1}}

\newcommand{\dpseg}{\texttt{dpseg}}
\newcommand{\remove}[1]{}

\title{Unwritten Languages Demand Attention Too! Word Discovery with Encoder-Decoder Models}

\name{Marcely Zanon Boito$^{1,2}$, Alexandre B\'erard$^1$, Aline Villavicencio$^2$ and Laurent Besacier$^1$}
\address{$^1$Laboratoire d'Informatique de Grenoble, Univ. Grenoble Alpes (UGA), France\\
$^2$ Institute of Informatics, UFRGS, Brazil}

\begin{document}
%
\maketitle
\begin{abstract}

Word discovery is the task of extracting words from unsegmented text. In this paper we examine to what extent neural networks can be applied to this task in a realistic unwritten language scenario, where only small corpora and limited annotations are available. We investigate two scenarios: one with no supervision and another with limited supervision with access to the most frequent words. Obtained results show that it is possible to retrieve at least 27\% of the gold standard vocabulary by training an encoder-decoder neural machine translation system with only 5,157 sentences. This result is close to those obtained with a task-specific Bayesian nonparametric model. Moreover, our approach has the advantage of generating translation alignments, which could be used to create a bilingual lexicon. As a future perspective, this approach is also well suited to work directly from speech.


\end{abstract} 
\begin{keywords}
Word Discovery, Computational Language Documentation, Neural Machine Translation, Attention models
\end{keywords}
\section{Introduction}

Computational Language Documentation (CLD) aims at creating tools and methodologies to help automate the extraction of lexical, morphological and syntactic information in languages of interest.
This paper focuses on languages (most of them endangered and unwritten) spoken in small communities all across the globe. Specialists believe that more than 50\% of them will become extinct by the year 2100~\cite{austin2011cambridge}, and
manually documenting all these languages is not feasible.
Initiatives for helping with this issue include organizing tasks~\cite{versteegh2016zero,jansen2013summary} and proposing pipelines for automatic information extraction from speech signals~\cite{besacier2006towards,bartels2016toward,bansal2016weakly,lignos2010recession,anastasopoulos2017case}.

Methodologies for CLD 
should consider the nature of the collected data: endangered languages may lack a well-defined written form (they often are oral-tradition languages). Therefore, in the absence of a standard written form, one alternative is to align collected speech to its translation in a well-documented language. Due to the challenge of finding bi-lingual speakers to help in this documentation process, the collected corpora usually are of small size.

One of the tasks involved in the documentation process is word segmentation.
It consists of, given an unsegmented input, finding the boundaries between word-like units.
This input can be a sequence of characters or phonemes, or even raw speech. Such a system can be very useful to linguists, helping them start the transcription and documentation process. For instance, a linguist can use the output of such a system as an initial vocabulary, and then manually validate the generated words.
Popular solutions for this task are Nonparametric Bayesian models ~\cite{goldwater2009bayesian,lee2015unsupervised,elsner2013joint,adams2015inducing,adamslearning} and, more recently, Neural Networks~\cite{bartels2016toward, anastasopoulos2017case, wang2016morphological}. 
The latter have also been used for related tasks such as speech translation \cite{berard2016listen, duong2016attentional} or unsupervised phoneme discovery \cite{franke2016phoneme}.

\textbf{Contribution.} 
This paper is the first attempt to leverage attentional encoder-decoder models for language documentation of a truly unwritten language. We show that it is possible, from very little data
, to perform unsupervised word discovery with a performance (F-score) only slightly lower than that of Nonparametric Bayesian 
models, known to perform very well on this task in limited data settings.
Moreover, our approach aligns symbols in the unknown language with words from a known language which, as a by-product, bootstraps a bilingual dictionary.
Therefore, in the remainder of this paper, we will use the term \textit{word discovery} (instead of \textit{word segmentation}), since our approach does not only find word boundaries but also aligns word segments to their translation in another language.

Another reason why we are interested in attentional encoder-decoder models, is that they can easily be modified to work directly from the speech signal, which is our ultimate goal.

\textbf{Approach.} In a nutshell, we train an attention-based Neural Machine Translation (NMT) model, and extract the
soft-alignment probability matrices generated by the attention mechanism.
These alignments are then post-processed to segment a sequence of symbols (or speech features) in an unknown language (Mboshi) into words.
We explore three improvements for our neural-based approach: alignment smoothing presented in~\cite{duong2016attentional}, vocabulary reduction discussed in~\cite{godard2016preliminary}, and Moses-like symmetrization 
of our soft-alignment probability matrices. We also propose to reverse the translation direction, translating from known language words to unknown language tokens. 
Lastly, we also study a semi-supervised scenario, where prior knowledge is available, by providing the 100 most frequent words to the system.

\textbf{Outline.} This paper is organized as follows: we present related work in Section~\ref{sec:related}, and the neural architecture, corpus, and our complete approach in Section~\ref{sec:methodology}. Experiments and their results are presented in Section~\ref{sec:unsupervised} and \ref{sec:semi}, and are followed by an analysis in Section~\ref{sec:analysis}.
We conclude our work with a discussion about possible future extensions in Section~\ref{sec:conclusion}.

\vspace{-0.4cm}
\section{Related Work}\label{sec:related}

Nonparametric Bayesian Models (NB models)~\cite{goldwater2007nonparametric,johnson2009improving} are statistical approaches that can be used for word segmentation and morphological analysis. 
Recent variants of these models are able to work directly with raw speech~\cite{lee2015unsupervised}, or with sentence-aligned translations~\cite{adams2015inducing}.
The major advantage of NB models for CLD 
is their robustness \remove{tackling} to small training sets. Recently, \cite{godard2016preliminary} achieved their best results on a subset (1200 sentences) of the same corpus we use in this work by using a NB model. Using the \dpseg{} system\footnote{Available at http://homepages.inf.ed.ac.uk/sgwater/resources.html.}~\cite{goldwater2009bayesian}, they retrieved 23.1\% of the total vocabulary (type recall), achieving a type F-score of 30.48\%.


Although NB models are well-established in the area of unsupervised word discovery, we wish to explore what neural-based approaches could add to the field.
In particular, attention-based encoder-decoder approaches have been very successful in Machine Translation~\cite{bahdanau2014neural}, and have shown promising results in End-to-End Speech Translation~\cite{berard2016listen, weiss2017sequence} (translation from raw speech, without any intermediate transcription).
This latter approach is especially interesting for language documentation, which often uses corpora made of audio recordings aligned with their translation in another language (no transcript in the source language).

While attention probability matrices offer accurate information about word soft-alignments in NMT systems~\cite{bahdanau2014neural, berard2016listen}, we investigate whether this is reproducible in scenarios with limited amounts of training data. That is because a notable drawback of neural-based models is their need of large amounts of training data \cite{KoehnK17}.
\remove{This data is needed to train the numerous network parameters, and this amount is usually not available in low-resource scenarios.}

We are aware of only one other work using an NMT system for unsupervised word discovery in a low-resource scenario.
This work \cite{duong2016attentional} used an 18,300 Spanish-English parallel corpus to emulate an endangered language corpus. Their approach for unsupervised word discovery is the most similar to ours.
However, we go one step further: we apply such a technique to a real language documentation scenario. We work with only five thousand sentences in an unwritten African language (Mboshi), as we believe that this is more representative of what linguists may encounter when documenting languages.




\section{Methodology}\label{sec:methodology}

\remove{
The main objective of this work is to investigate the usefulness of encoder-decoder NMT pipeline for language documentation. A by-product of the pipeline is an alignment model that provides soft-alignment information that we use for performing 
word discovery.

However, neural approaches are known for needing huge amounts of data in order to train their uncountable parameters. Because of that, we focus on investigating if these approaches are robust to low-resource scenarios, where the data is in the order of several thousand of sentences only. 
}


\subsection{Mboshi-French Parallel Corpus}\label{subsec:corpus}

We use a 5,157 sentence parallel corpus in Mboshi (Bantu C25), an unwritten\footnote{Even though it is unwritten, linguists provided a non-standard grapheme form, considered to be close to the language phonology.} African language, aligned to French translations at the sentence level. Mboshi is a language spoken in Congo-Brazzaville, and it has 32 different phonemes (25 consonants and 7 vowels) and two tones (high and low). The corpus was recorded using the LIG-AIKUMA tool~\cite{blachon2016parallel} in the scope of the BULB project \cite{adda2016breaking}.

For each sentence, we have a non-standard grapheme transcription (the gold standard for segmentation), an unsegmented version of this transcription, a translation in French, 
a lemmatization\footnote{For tokenization and lemmatization we used TreeTagger~\cite{schmid2013probabilistic}.} of this translation, and an audio file. 
It is important to mention that in this work, we use Mboshi unsegmented non-standard grapheme form (close to language phonology) as a source while direct use of speech signal is left for future work.

We split the corpus into training and development sets, 
using 10\% for the latter.
Table \ref{tableDevTrain} gives a summary of the  types (unique words) and tokens (total word counts) on each side of the parallel corpus.



\begin{table}[]
\centering
\begin{tabular}{c|c|c|c|}
\cline{2-4}
\textbf{}                          & \textbf{\# types} & \textbf{\#tokens} & \textbf{\begin{tabular}[c]{@{}c@{}}avg \# tokens \\ per sentence\end{tabular}} \\ \hline
\multicolumn{1}{|c|}{Mboshi Dev}   & 1,324             & 3,133             & 6.0                                                                            \\ \hline
\multicolumn{1}{|c|}{Mboshi Train} & 6,245             & 27,579            & 5.9                                                                            \\ \hline
\multicolumn{1}{|c|}{French Dev}   & 1,343             & 4,321             & 8.2                                                                            \\ \hline
\multicolumn{1}{|l|}{French Train} & 4,903             & 38,226            & 8.4                                                                            \\ \hline
\end{tabular}
\caption{Organization of the corpus in development (Dev, 514 sentences) and training (Train, 4,643 sentences) sets for the neural model.}
\label{tableDevTrain}
\end{table}

\subsection{Neural Architecture}\label{subsec:model}

We use the LIG-CRIStAL NMT system\footnote{Available at https://github.com/eske/seq2seq.}\remove{with the global attention mechanism presented in~\cite{bahdanau2014neural}}, using unsegmented text input for training. The model is easily extendable to work directly with speech~\cite{berard2016listen}.
Our NMT models follow \cite{bahdanau2014neural}. A bidirectional encoder reads the input sequence $x_1,...,x_A$ and produces a sequence of
encoder states $\mathbf{h} = h_1,...,h_A  \in \mathbb{R}^{2\times n}$, where $n$ is the chosen encoder cell size. A decoder uses its current state $s_t$ and an attention mechanism to generate the next output symbol $z_t$.
At each time step $t$, the decoder computes a probability distribution over the target vocabulary. Then, it generates the symbol $z_t$ whose probability is the highest (it stops once it has generated a special end-of-sentence symbol).
The decoder then updates its state $s_t$ with the generated token $z_t$. In our task, since reference translations are always available (even at test time), we always force feed  previous ground-truth symbol $w_t$ instead of the generated symbol $z_t$ (teacher forcing).
\begin{align}
    c_t={\rm attn}(\mathbf{h}, s_{t-1}) \\
    y_t={\rm output}(s_{t-1} \oplus E(w_{t-1}) \oplus c_t) \\
    z_t=\arg \max{y_t} \\
    s_t={\rm LSTM}(s_{t-1}, E(w_t) \oplus c_t)
\end{align}

\noindent $\oplus$ is the concatenation operator. $s_0$ is initialized with the last state of the encoder (after a non-linear transformation), $z_0 = \texttt{<BOS>}$ (special token), and $E\in\mathbb{R}^{|V|\times n}$ is the target embedding matrix.
The $output$ function uses a maxout layer, followed by a linear projection to the vocabulary size $|V|$.

The attention function is defined as follows:

\vspace{-0.5cm}
\begin{align}
    \label{eq:global}
    c_t = {\rm attn}(\mathbf{h}, s_t) = \sum_{i=1}^{A}\alpha_i^t h_i \\
    \alpha_i^t = {\rm softmax}(e_i^t) \label{eq:attn} \\
    e_i^t = v^T\tanh{(W_1h_i+W_2s_t + b_2)}
\end{align}
\noindent where $v$, $W_1$, $W_2$, and $b_2$ are learned jointly with the other parameters of the model.
At each time step ($t$) a score $e_i^t$ is computed for each encoder state $h_i$, using the current decoder state $s_t$. These scores are then normalized 
using a $softmax$ function, thus giving a probability distribution over the input sequence $\sum_{i=1}^{A}{\alpha_i^t} = 1$ and $\forall{i}, 0\leq \alpha_i^t \leq1$.
The context vector $c_t$ used by the decoder, is a weighted sum of the encoder states. This can be understood as a summary of the useful information in the input sequence for the generation of the next output symbol $z_t$.
The weights $\alpha^t_i$ can be seen as a soft-alignment between input $x_i$ and output $z_t$.

Our models are trained using the Adam algorithm, with a learning rate of $0.001$ and batch size ($N$) of $32$. We minimize a cross-entropy loss between the output probability distribution $p_t=softmax(y_t)$ and reference translation $w_t$:

\vspace{-0.5cm}
\begin{align}
    L = \frac{1}{N}\sum_{i=1}^{N}{{\rm loss}(s_i=w_1,...,w_T\mid\mathbf{x_i})} \\
    {\rm loss}(w_1,..,.w_T\mid\mathbf{x_i}) = -\sum_t^T{\sum_j^{|V|}{\log{p_{tj}}\times \mathbbm{1}(w_t=V_j) }} \\
    p_{tj} = \frac{e^{y_{tj}}} { \sum_k^{|V|}{e^{y_{tk}}} }
\end{align}

\subsection{Neural Word Discovery Approach}\label{subsec:approach}

\begin{figure}
\centering
\includegraphics[scale=0.35]{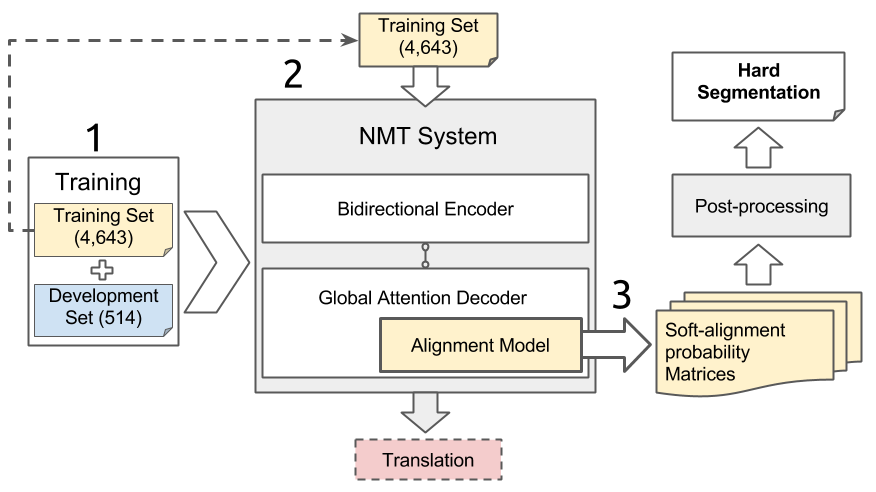}
\caption{Neural word discovery pipeline.}
\label{transductiveschema}
\end{figure}

Our full word discovery pipeline is illustrated in Figure~\ref{transductiveschema}. We start by training\remove{the presented encoder-decoder} an NMT system using the Mboshi-French parallel corpus, without the word boundaries on the Mboshi side. This is shown as step 1 in the figure. 

We stop training once the training loss stops decreasing. At this point, we expect the alignment model to be the most accurate on the training data.
Then we ask the model to force-decode the entire training set.
We extract soft-alignment probability matrices computed by the attention model while decoding (step 2).


Finally, we post-process this soft-alignment information and infer a word segmentation (step 3). We first transform the soft-alignment into a hard-alignment, by aligning each source symbol $x_i$ with target word $w_t$ such that: $t=\arg\max_{i} {\alpha_i^t}$.
Then we segment the input (Mboshi) sequence according to these hard-alignments: if two consecutive symbols are aligned with the same French word, they are considered to belong to the same Mboshi word.

\section{Unsupervised Word Discovery Experiments}\label{sec:unsupervised}

For the unsupervised word discovery experiments, we used the unsegmented transcription in Mboshi provided by linguists, aligned with French sentences. This Mboshi unsegmented transcription is made of 44 different symbols. 

We experimented with the following variations:

\begin{table*}[]
\centering
\begin{tabular}{c|c|c|c|c|c|c|}
\cline{2-7}
\multicolumn{1}{l|}{}                                                                                          & \multicolumn{3}{c|}{\textbf{TOKENS}}                     & \multicolumn{3}{c|}{\textbf{TYPES}}                    \\ \cline{2-7} 
\textbf{}                                                                                                      & \textbf{Recall} & \textbf{Precision} & \textbf{F-score} & \textbf{Recall} & \textbf{Precision} & \textbf{F-score} \\ \hline
\multicolumn{1}{|c|}{Base Model (Mb-Fr)}                                                                       & 7.16            & 4.50               & 5.53             & 12.85           & 6.41               & 8.55             \\ \hline
\multicolumn{1}{|c|}{\begin{tabular}[c]{@{}c@{}}Base Model (Mb-Fr)\\ with Alignment Smoothing\end{tabular}}    & 6.82            & 5.85               & 6.30             & 15.00           & 6.76               & 9.32             \\ \hline
\multicolumn{1}{|c|}{Reverse Model (Fr-Mb)}                                                                    & 20.04           & 10.02              & 13.36            & 18.62           & 14.80              & 16.49            \\ \hline
\multicolumn{1}{|c|}{\begin{tabular}[c]{@{}c@{}}Reverse Model (Fr-Mb)\\ with Alignment Smoothing\end{tabular}} & 21.44           & 16.49              & 18.64            & 27.23           & 15.02              & 19.36            \\ \hline
\end{tabular}
\caption{Unsupervised Word Discovery results with 4,643 sentences.\remove{alex: explain tokens and types}}
\label{resultsUnsupervised}
\end{table*}

\begin{enumerate}
    \item \textbf{Alignment Smoothing}: to deal with source (phones or
graphemes) vs.\ target (words) sequence length discrepancy, we need to encourage many-to-one alignments between Mboshi and French. These alignments are needed in order to cluster Mboshi symbols into word-units. 
For this purpose,
we implemented the alignment smoothing proposed by \cite{duong2016attentional}.
The softmax function used by the attention mechanism (see eq.~\ref{eq:attn}) takes an additional \emph{temperature} parameter: $\alpha_i^t=\exp{(e_i^t/T)}/\sum_j{\exp{(e_j^t/T)}}$
A temperature $T$ greater than one\footnote{\marcely{We use $T=10$, like the original paper \cite{duong2016attentional}.}} will result in a less sharp softmax, which boosts many-to-one alignments. 
    In addition, the probabilities are smoothed by averaging each score with the scores of the two neighboring words: $\alpha^t_i\leftarrow(\alpha^t_{i-1}+\alpha^t_{i}+\alpha^t_{i+1})/3$ (equivalent to a low-pass filtering on the soft-alignment probability matrix).

    \item \textbf{Reverse Architecture}: in NMT, the soft-alignments are created by forcing the probabilities for each target word $t$ to sum to one (i.e. $\sum_i \alpha_i^t = 1$). However, there is no similar constraint for the source symbols, as discussed in~\cite{duong2016attentional}.
    Because we are more interested in the alignment than the translation itself,  we propose to reverse the architecture. The reverse model translates from French words to Mboshi symbols. This prevents the attention model from ignoring some Mboshi symbols.
    
    

    \item \textbf{Alignment Fusion}: statistical machine translation systems, such as the Moses~\cite{koehn2007moses}, extract alignments in both directions (source-to-target and target-to-source) and then merge them, creating the final translation model. This alignment fusion is often called symmetrization.
    We investigate whether this Moses-like symmetrization improves our results by merging 
    the soft-alignments probability matrices generated by our base (Mboshi-French) and reverse (French-Mboshi) models. We replace each probability $\alpha_i^t$ by  $\frac{1}{2} (\alpha_i^t + \beta_t^i)$, where  $\beta_t^i$ is the probability for the same alignment $i\leftrightarrow t$ in the reverse architecture.
    
    \item \textbf{Target Language Vocabulary Reduction}: to reduce vocabulary size on the known language, we replace French words by their lemmas. The intuition is that, by simplifying the translation information, the model could more easily learn relations between the two languages. For the task of unsupervised word discovery, this technique was recently investigated by~\cite{godard2016preliminary}.

\end{enumerate}

The base model (Mboshi to French) uses an embedding size and cell size of 12. The encoder stacks two bidirectional LSTM layers, and the decoder uses a single LSTM layer.
The reverse model (French to Mboshi) uses an embedding size and cell size of 64, with a single layer bidirectional encoder and single layer decoder. 

We present in Table~\ref{resultsUnsupervised} the unsupervised word discovery task results obtained with our base model, and with the reverse model, with and without alignment smoothing (items 1 and 2). We notice that the alignment smoothing technique presented by~\cite{duong2016attentional} improved the results, especially for types.

Moreover, we show that the proposed reverse model considerably improves type and token retrieval.
This seems to confirm the hypothesis that reversing the alignment direction results in a better segmentation (because the attention model has to align each Mboshi symbol to French words with a total probability of 1).
This may also be due to the fact that the reverse model reads words and outputs character-like symbols which is generally easier than reading sequences of characters \cite{LeeCH16}. 
Finally, we achieved our best result by using the reverse model with alignment smoothing (last row in Table~\ref{resultsUnsupervised}).

We then used this latter model for testing alignment fusion and vocabulary reduction (items 3 and 4).
For alignment fusion, we tested three configurations using matrices generated by the base and reverse models. We tested the fusion of the raw soft-alignment probability matrices (without alignment smoothing), the fusion of already smoothed matrices, as well as this latter fusion followed by a second step of smoothing. All these configurations lead to negative results: recall reduction between 3\% and 5\% for tokens and between 1\% and 9\% for types. 
We believe this happens because by averaging the reverse model's alignments with the ones produced by the base model (which does not have the constraint of using all the symbols) we degrade the generated alignments, more than exploiting information discovered in both directions.

Lastly, when running the reverse architecture (with alignment smoothing) using French lemmas (vocabulary  reduction), we also noticed a reduction in performance. The lemmatized model version had a recall drop of approximately 2\% for all tokens and types metrics. We believe this result could be due to the nature of the Mboshi language, and not necessarily a generalizable result. Mboshi has a rich morphology, creating a different word for each verb tense, which includes radical and all tense information. Therefore, by removing this from the French translations, we may actually make the task harder, since the system is forced to learn to align different words in Mboshi to the same word in French.

\section{Semi-supervised Word Discovery Experiments}\label{sec:semi}

A language documentation task is rarely totally unsupervised, since linguists usually immerse themselves in the community when documenting its language. In this section, we explore a semi-supervised approach for word segmentation, using our best reverse model from Section~\ref{sec:unsupervised}.

To emulate prior knowledge, we select the 100 most frequent words in the gold standard for Mboshi segmentation. We consider this amount reasonable for representing the information a linguist could acquire after 
a few days. Our intuition is that providing the segmentation for these words could help improve the performance of the system for the rest of the vocabulary. 

To incorporate this prior information to our system, we simply add known tokens on the Mboshi side of the corpus, keeping the remaining symbols unsegmented. This creates a mixed representation, in which the Mboshi input has at the same time unsegmented symbols and segmented words. Since languages follow Zipfian distributions~\cite{powers1998applications} and we are giving to the model the most frequent words in the corpus, analysis is not done in terms of tokens, since this would be over-optimistic and bias the model evaluation, but only in terms of types. Results are presented in Table~\ref{dictResults}.
   
\begin{table}
\centering
\begin{tabular}{c|c|c|}
\cline{2-3}
\textbf{}                                         & \textbf{Unsupervised} & \textbf{Semi-supervised} \\ \hline
\multicolumn{1}{|c|}{\textbf{Recall}}             & 27.23                 & 29.49                    \\ \hline
\multicolumn{1}{|c|}{\textbf{Precision}}          & 15.02                 & 24.64                    \\ \hline
\multicolumn{1}{|c|}{\textbf{F-score}}            & 19.36                 & 26.85                    \\ \hline
\multicolumn{1}{|c|}{\textbf{\# correct types}}   & 1,692                 & 1,842                    \\ \hline
\multicolumn{1}{|c|}{\textbf{\# generated types}} & 11,266                & 7,473                    \\ \hline
\end{tabular}
\caption{Types results for the semi-supervised word discovery task (100 known words, 4.653 sentences).}
\label{dictResults}
\end{table}

For types, we observed an increase of 2.4\% in recall. This is not a huge improvement, considering that we are giving 100 words to the model. We discovered that our unsupervised model was already able to discover 97 of these 100 frequent words, which could justify the small performance difference between the models. 
In addition to the 100 types already known, the semi-supervised model found 50 new types that the unsupervised system was unable to discover.

Finally, it is interesting to notice that, while the performance increase is not huge, the semi-supervised system reduced considerably the number of types generated, from 11,266 to 7,473. This suggests that this additional information helped the model to create a better vocabulary representation, closer to the gold standard vocabulary.

\section{Analysis}\label{sec:analysis}


\subsection{Baseline Comparison}\label{subsec:baseline}

As a 
baseline, we used \dpseg{} 
 \cite{Goldwater06contextual,Goldwater09bayesian} which implements a Nonparametric Bayesian approach, where (pseudo)-words are generated by a bigram model over a non-finite inventory, through the use of a Dirichlet-Process.

We used the same hyper-parameters 
as \cite{godard2016preliminary}, which were tuned on a larger English corpus and then successfully applied to the segmentation of Mboshi. We use a random initialization and 19,600 sampling iterations.


Table~\ref{dpsegandsu} shows our results for types compared to the NB model. Although the former 
is able to retrieve more from the vocabulary,  the latter 
has higher precision, and both  are close in terms of F-score. Additionally, ours has the advantage of providing clues for translation.

\begin{table}[]
\centering
\begin{tabular}{c|c|c|c|c|}
\cline{2-5}
\textbf{}                                                                                                               & \textbf{Recall} & \textbf{Precision} & \textbf{F-score} & $\sigma$ \\ \hline
\multicolumn{1}{|c|}{\textbf{\begin{tabular}[c]{@{}c@{}}Reverse Model \\ (Fr-Mb) with AS\end{tabular}}} & 27.23           & 15.02              & 19.36            & 0.032  \\ \hline
\multicolumn{1}{|c|}{\textbf{dpseg}}                                                                                    & 13.94           & 38.32              & 20.45            & 0.272   \\ \hline
\end{tabular}
\caption{Comparison between the NB model (dpseg) and the reverse model with alignment smoothing (AS) for unsupervised word discovery.
The scores were obtained by averaging over three instances of each model.
}
\label{dpsegandsu}
\end{table}

It is interesting to notice that our neural approach, which is not specialized for this task (the soft-alignment scores are only a by-product of translation), was able to achieve close performance to the \dpseg{} method, which is known to be very good in low-resource scenarios. This highlights the potential of our approach for language documentation.

\subsection{Vocabulary Analysis}\label{subsec:vocabulary}

\begin{figure}
\centering
\includegraphics[scale=0.45]{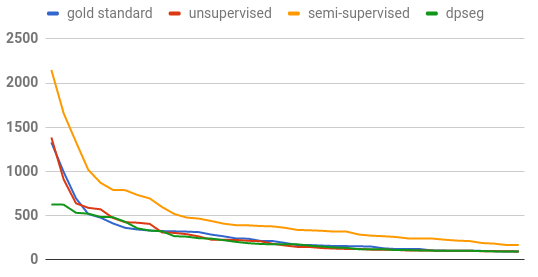}
\caption{Word frequency distribution of the three models and the gold standard distribution.}
\label{zipfdist}
\end{figure}

To understand the segmentation behavior of our approach, we looked at the generated vocabulary. We compare our unsupervised and semi-supervised methods with the gold standard and the NB baseline, \dpseg{}. The first characteristic we looked at was the word distribution of the generated vocabularies.
While we already knew that \dpseg{} constraints the generated vocabulary to follow a power law, we observed that our approaches also display such a behavior. They produce curves that are as close to the real language distribution as \dpseg{} (see Figure~\ref{zipfdist}).

\begin{figure}
\centering
\includegraphics[scale=0.42]{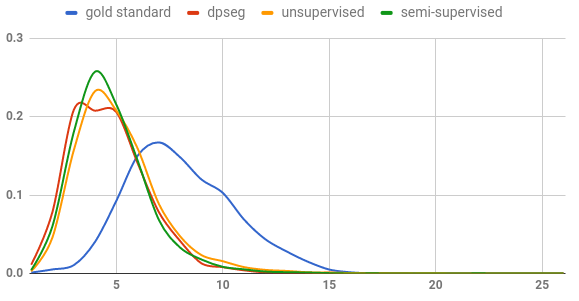}
\caption{Type length distribution of the gold standard, \dpseg{} and our unsupervised and semi-supervised methods.}
\label{typesdist}
\end{figure}

We also measured the average word length to identify under-segmentation and over-segmentation. To be able to compare vocabularies of varying sizes, we normalized the frequencies by the total number of generated types. The curves are shown in Figure~\ref{typesdist}. Reading the legend from left to right, the vocabulary sizes are 6,245, 2,285, 11,266, and 7,473.


Our semi-supervised configuration is the closest to the real vocabulary in terms of vocabulary size, with only 1,228 more types.
All the approaches (including \dpseg{}) over-segment the input in a similar way, creating vocabularies with average word length of four (Figure~\ref{typesdist}).

Since both \dpseg{} and neural-based approaches suffer from the same over-segmentation problem, we believe that this is a consequence of the corpus used for training, and not necessarily a general characteristic of our approach in low-resource scenarios.
For our neural approaches, another justification is the corpus being small, and the average tokens per sentence being higher at the French side (shown in Table~\ref{tableDevTrain}), which can potentially disperse the alignments over the possible translations, creating multiple boundaries.

\begin{figure}
\centering
\includegraphics[width=.85\linewidth]{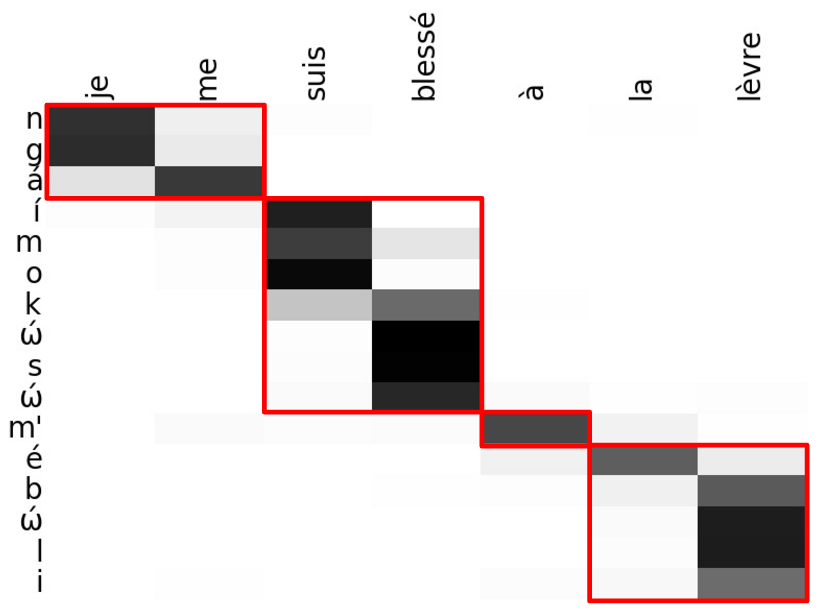}
\caption{Example of soft-alignment \remove{probability matrix from} generated by our unsupervised word discovery model. The darker the square, the higher is the probability for the source-target pair. Our segmentation was ``ng\'a \'imo k$\acute{\omega}$s$\acute{\omega}$ m' \'e b$\acute{\omega}$li'', while the correct one is ``ng\'a \'imok$\acute{\omega}$s$\acute{\omega}$ m' \'eb$\acute{\omega}$li''.}
\label{fig:heatmap}
\end{figure}

Moreover, as Mboshi is an agglutinative language, 
there were several cases in which we had a good alignment but wrong segmentation. An example is shown in Figure~\ref{fig:heatmap}, where we see that the word ``\'imok$\acute{\omega}$s$\acute{\omega}$'' was split in two words in order to keep its alignment to both parts of its French translation ``suis bless\'e''. This is also the case of the last word in this figure: Mboshi does not require articles preceding nouns, which caused misalignment. We believe that by exploiting translation alignment, we could constraint our segmentation procedure, creating a more accurate word discovery model. Finally, we were able to create a model of reasonable quality which gives segmentation and alignment information using only 5,157 sentences for training (low-resource scenario).


\section{Conclusion}\label{sec:conclusion}

In this work, we presented a neural-based approach for performing word discovery in low-resource scenarios. We used an NMT system with global attention to retrieve soft-alignment probability matrices between source and target language, and we used this information to segment the language to be documented. 
A similar approach was presented in \cite{duong2016attentional}, but this work represents the first attempt at training a neural model with a real unwritten language based on a small corpus made of only 5,157 sentences.

By reversing the system's input order and applying alignment smoothing, we were able to retrieve 27.23\% of the vocabulary, which gave us an F-score close to the NB baseline, known for being robust to low-resource scenarios. Moreover, this approach has the advantage of naturally incorporating translation, which can be used for enhancing segmentation and creating a bilingual lexicon. The system is also easily extendable to work with speech, a requirement for most of the approaches in CLD.

Finally, as future work, our objective is to discover lexicon directly from speech, inspired by the encoder-decoder architectures presented in \cite{berard2016listen, weiss2017sequence}. We will also explore different training objective functions more correlated with segmentation quality, in addition to MT metrics. Lastly, we intend to investigate more sophisticated segmentation methods from the generated soft-alignment probability matrices, identifying the strongest alignments in the matrices, and using their segmentation as prior information to the system (iterative segmentation-alignment process).

\bibliographystyle{IEEEbib}
\bibliography{refs}

\end{document}